\documentclass[journal]{IEEEtran} 

\usepackage{cite} 
\usepackage{amsmath,amssymb,amsfonts}
\usepackage{graphicx}
\usepackage{booktabs}
\usepackage{multirow}
\usepackage{array}
\usepackage{tabularx}
\usepackage{longtable}
\usepackage[table]{xcolor}
\usepackage{url}
\usepackage{xcolor}
\usepackage[linesnumbered,ruled,vlined]{algorithm2e}

\usepackage[caption=false,font=footnotesize]{subfig}

\usepackage[hidelinks]{hyperref}

\newenvironment{linenomath}{}{}
\newenvironment{linenomath*}{}{}

\title{An Intention-Driven Lane Change Framework Considering Heterogeneous Dynamic Cooperation in Mixed-Traffic Environments}

\author{
  Xiaoyun~Qiu,
  Haichao~Liu,
  Yue~Pan,
  Jun~Ma,
  and~Xinhu~Zheng*\thanks{*Corresponding author}
  \thanks{X. Qiu, Y. Pan, J. Ma, and X. Zheng are with the Intelligent Transportation Thrust, Systems Hub, The Hong Kong University of Science and Technology (Guangzhou), Guangzhou, China, and with the Guangdong Provincial Key Lab of Integrated Communication, Sensing and Computation for Ubiquitous Internet of Things, Guangzhou, China.}%
  \thanks{H. Liu and J. Ma are with the Robotics and Autonomous Systems Thrust, Systems Hub, The Hong Kong University of Science and Technology (Guangzhou), Guangzhou, China.}%
}

\begin{document}
\maketitle

\begin{abstract}
In mixed-traffic environments, autonomous vehicles (AVs) must interact with heterogeneous human-driven vehicles (HVs) whose intentions and driving styles vary across individuals and scenarios. Such variability introduces uncertainty into lane change interactions, where safety and efficiency critically depend on accurately anticipating surrounding drivers’ cooperative responses. Existing methods often oversimplify these interactions by assuming uniform or fixed behavioral patterns.  
To address this limitation, we propose an \emph{intention-driven lane change framework} that integrates driving-style recognition with cooperation-aware decision-making and motion-planning. A deep learning-based classifier identifies distinct human driving styles in real time. We then introduce a dual-perspective cooperation score composed of intrinsic style-dependent tendencies and interactive dynamic components, enabling interpretable and adaptive intention prediction and quantitative inference. A decision-making module combines behavior cloning (BC) and inverse reinforcement learning (IRL) to determine lane change feasibility. Later, a coordinated motion-planning architecture integrating IRL-based intention inference with model predictive control (MPC) is established to generate collision-free and socially compliant trajectories.  
Experiments on the NGSIM dataset show that the proposed decision-making model outperforms representative rule-based and learning-based baselines, achieving 96.98\% accuracy in lane change classification. Motion-planning evaluations further demonstrate improved maneuver success and execution stability in mixed-traffic environments.
These results validate the effectiveness of structured cooperation modeling for intention-driven autonomous lane changes.

\end{abstract}


\section{Introduction}

The rapid advancement of autonomous driving technology has accelerated the transition toward mixed-traffic environments where autonomous vehicles (AVs) will inevitably share roadways with human-driven vehicles (HVs) for the foreseeable future~\cite{yurtsever2020survey}. Within this operational paradigm, the ability of AVs to accurately interpret and adapt to human drivers' heterogeneous behavioral patterns during lane change maneuvers emerges as a critical safety imperative~\cite{xing2021toward, chen2022milestones}. This challenge is particularly acute given the fundamental uncertainty inherent in human decision-making processes and the dynamic complexity of traffic interactions~\cite{schwarting2018planning}.

Modern lane change algorithms have evolved from classical models~\cite{gipps1986model} to contemporary data-driven approaches~\cite{zheng2014recent, wang2019review}, and can be broadly categorized into three paradigms: rule-based, data-driven, and incentive-based systems. Recent advances in inverse reinforcement learning~\cite{sun2022inverse} and probabilistic modeling~\cite{sun2018probabilistic} have partially incorporated driver intentions into decision-making~\cite{sheng2022cooperation}. Nevertheless, several critical limitations persist. Firstly, existing frameworks often rely on a behavioral homogeneity assumption, which oversimplifies surrounding human drivers by adopting uniform behavioral models and thereby overlooks inter-driver heterogeneity in decision-making. Secondly, although some methods incorporate personalized intentions, they typically fail to capture the temporal evolution of interactions among surrounding vehicles. Lastly, there remains an intention modeling gap: current prediction approaches lack mechanisms to effectively link observable driving styles with latent reward structures, resulting in a disconnect between behavior recognition and motion-planning.

These limitations are particularly problematic given that sudden lane changes account for about 17.0\% of total severe crashes~\cite{shawky2020factors}, underscoring the urgent need for AV frameworks capable of simultaneously predicting trajectories and adapting to individual driving styles~\cite{drivingstyle}.

\begin{figure*}[!t]
  \centering
  \includegraphics[width=1\textwidth]{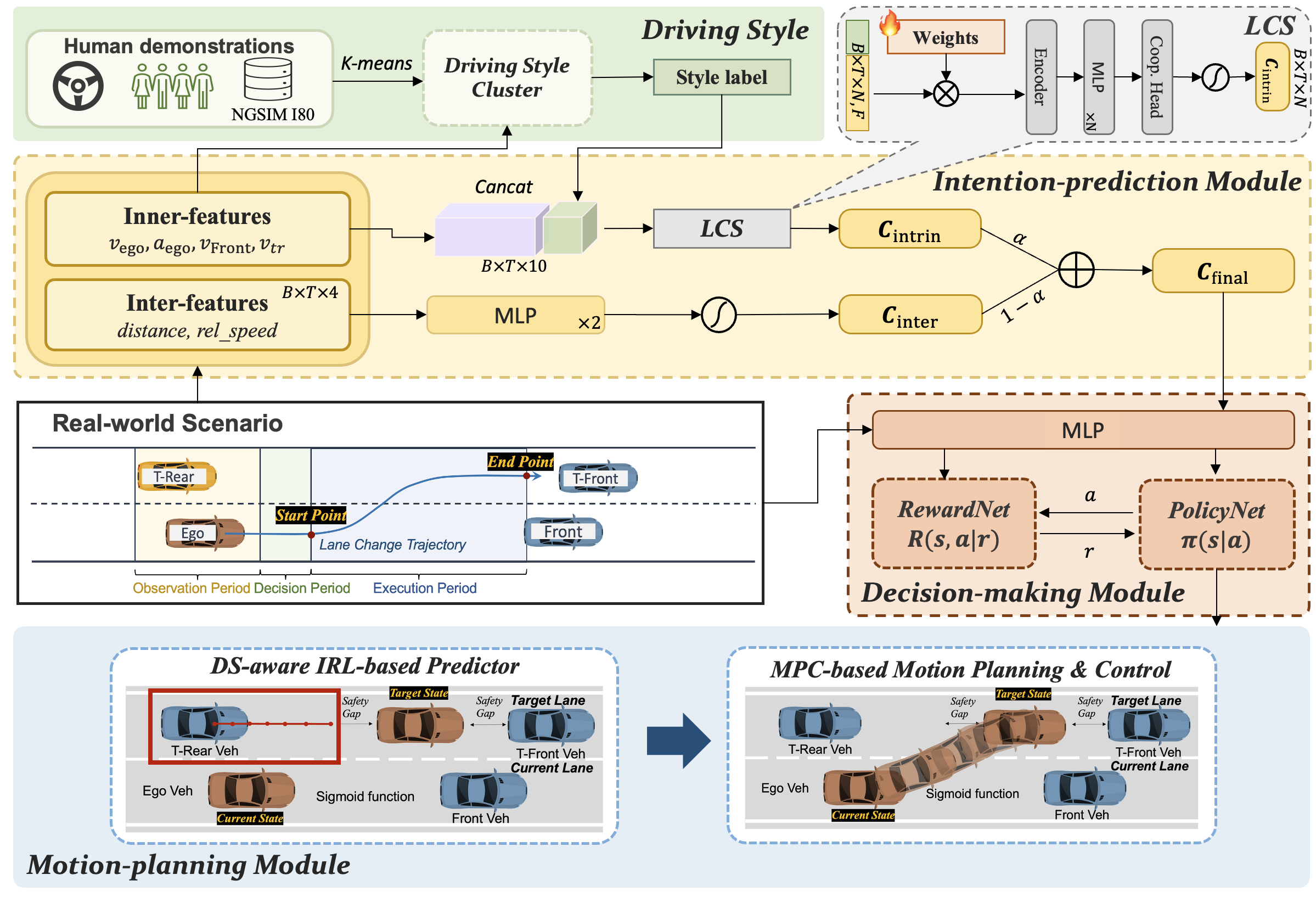}
  \caption{Overall architecture of the proposed intention-driven lane change framework. 
(i) \textbf{Driving-style recognition} provides discrete style labels; 
(ii) \textbf{Intention prediction} estimates cooperation levels via intrinsic learnable cooperation score (LCS) and interactive dynamic cooperation score (DCS), fused into $c_{\text{final}}$; 
(iii) \textbf{Decision-making} integrates $c_{\text{final}}$ with vehicle states to predict lane change intentions using policy and reward networks under a BC–IRL training objective; 
(iv) \textbf{Motion-planning} incorporates IRL-based trajectory prediction of surrounding vehicles and an MPC-based controller to generate safe, executable lane change trajectories.}
  \label{fig:framework}
\end{figure*}

To address the identified shortcomings in existing algorithms, we introduce an intention-driven lane change framework that synergistically integrates driving-style recognition with adaptive decision-making as shown in Fig.\ref{fig:framework}. In the context of this research, ``driving style" is conceptualized as a habitual manner of vehicle operation, distinctive of an individual or subset of drivers, that serves as a proxy for underlying personalized driving intentions. Our proposed lane change framework incorporates driving style into two distinct yet interrelated components: intention-prediction and motion-planning. The main contributions of this study are summarized as follows:
\begin{itemize}
    \item We propose an explicit cooperation modeling mechanism for lane change interactions that quantifies heterogeneous human drivers’ cooperative intentions through a unified cooperation score, rather than assuming homogeneous behavioral priors.
    \item We introduce a dual-perspective cooperation decomposition that combines intrinsic style-dependent tendencies with interaction-driven dynamics, enabling interpretable and adaptive intention prediction of surrounding human drivers.
    \item We develop a BC-IRL decision-making framework in which the learned cooperation representation regularizes policy learning, leading to human-like lane change decisions that balance imitation fidelity and interaction adaptability.
    \item We propose an IRL-MPC pipeline that jointly learns style-specific reward functions and generates collision-free, kinematically feasible trajectories, enabling safe and personalized maneuver planning in mixed-traffic with heterogeneous human behaviors.
\end{itemize}

\section{Related Work}
\subsection{Automated Lane Change Algorithms}

Lane change algorithms have evolved from Gipps’ seminal car-following model~\cite{gipps1986model} into three methodological paradigms: rule-based, data-driven, and incentive-based approaches~\cite{zheng2014recent, wang2019review}. These paradigms reflect the shift from handcrafted heuristics toward learning- and optimization-driven strategies, yet each exhibits inherent strengths and weaknesses.  

\textbf{Rule-based methods} rely on predefined criteria to reproduce lane change behaviors. Representative studies include MOBIL~\cite{kesting2007general}, which minimizes overall braking induced by lane changes, and fuzzy-rule systems combined with LSTMs for maneuver forecasting~\cite{wang2021intelligent}. Rule-based constraints have also been integrated into longitudinal control~\cite{wang2019lane} and optimization formulations~\cite{he2021rule}. Although interpretable and computationally efficient, such approaches are rigid and sensitive to rule specification, often resulting in ambiguous decisions under complex traffic conditions.  

\textbf{Data-driven methods} exploit large-scale trajectory datasets to learn nonlinear mappings between traffic states and lane change decisions. Classical examples include SVM-based classifiers~\cite{mandalia2005using} and ANN–SVM hybrids~\cite{dou2016lane}, while deep learning has enabled richer feature representation through DNNs~\cite{liu2019deep}. Recent work further leverages Transformer architectures to capture temporal dependencies~\cite{guo2025vehicle}. Despite improved accuracy, data-driven models often lack interpretability and generalizability, making their deployment in heterogeneous traffic environments non-trivial.  

\textbf{Incentive-based methods} formulate lane change behavior as a sequential decision-making problem guided by rewards. Reinforcement learning has demonstrated promising results in adaptive driving policies~\cite{hoel2019combining, zhu2018human}, but suffers from reward design difficulties and sparse exploration. Inverse reinforcement learning (IRL) alleviates this by recovering latent reward structures from demonstrations~\cite{arora2021survey, sun2022inverse, huang2021driving}, enabling more human-like behavior synthesis. Nevertheless, IRL approaches remain computationally demanding and limited by state-space coverage.  

In summary, rule-based methods offer interpretability but are limited by rigid assumptions, data-driven approaches achieve higher predictive accuracy but often lack interpretability and robustness, while incentive-based methods capture sequential decision-making but face challenges in reward design and scalability. Although both academic and industrial systems have demonstrated safe and efficient lane change performance in constrained domains, existing frameworks still struggle to accommodate heterogeneous and dynamically evolving human behaviors. Motivated by these limitations, we leverage the complementary strengths of data-based learning (for accurate pattern recognition) and incentive-based reasoning (for human-like reward modeling) to design an intention-driven lane change framework that explicitly incorporates human drivers’ cooperation intentions into the decision-making process.

\subsection{Lane Change Frameworks Considering Intention Prediction}

To enhance safety and efficiency, recent lane change frameworks increasingly incorporate the prediction of surrounding HVs intentions. A number of studies have integrated intention prediction into decision-making pipelines~\cite{sun2022inverse, sun2018probabilistic, sheng2022cooperation}.  

Sun \textit{et al.}~\cite{sun2018probabilistic} proposed a hierarchical decision-making model in which the first layer interacts with nearby vehicles to generate preliminary lane change decisions, while the second layer refines these into detailed trajectories. Guo \textit{et al.}~\cite{guo2019improved} improved adaptive cruise control by embedding merging-behavior prediction into car-following dynamics, thereby enhancing comfort and safety. Leveraging deep sequence models, Gao \textit{et al.}~\cite{gao2023dual} developed a dual-Transformer architecture that jointly predicts surrounding HVs’ intentions and trajectories, demonstrating high accuracy by explicitly modeling social interactions. Zhao \textit{et al.}~\cite{zhao2023lc} emphasized intention prediction within responsibility-sensitive safety frameworks, highlighting its role in proactive risk mitigation. More recently, Wang \textit{et al.}~\cite{wang2025socially} introduced a socially aware DAG-based intention-prediction algorithm, which further improves reliability by capturing structural dependencies among surrounding vehicles.  

Despite these advances, most intention-aware frameworks still adopt simplifying assumptions. In particular, they often treat surrounding drivers as behaviorally homogeneous or rely on static personality traits, neglecting the dynamic interplay between driver heterogeneity and evolving traffic contexts. This limitation motivates approaches that explicitly integrate personalized and interactive cooperation intentions into lane change decision-making.

\subsection{Human Personality Integrated Lane Change Frameworks}

In mixed-traffic environments, frameworks that integrate human personality factors offer more adaptive and realistic decision-making for AVs. Human factors have therefore been incorporated into both decision-making and motion-planning modules. Sheng \textit{et al.}~\cite{sheng2022cooperation} designed a human-factor evaluation index emphasizing safety, efficiency, and comfort, which was subsequently embedded into a motion-planning algorithm that coordinated AV decisions with the interactive behaviors of surrounding vehicles. Liu \textit{et al.}~\cite{liu2019novel} incorporated benefit, safety, and tolerance factors into a support vector machine with Bayesian optimization to improve lane change rationality. Personalized driver traits have also been modeled explicitly: \cite{butakov2014personalized}, \cite{liu2022inverse}, and \cite{rosbach2019driving} developed strategies that adapt lane change policies to driver-specific characteristics. Zhu \textit{et al.}~\cite{zhu2018personalized} proposed a personalized assistance system that prioritizes safety by classifying drivers through a backpropagation neural network optimized with particle swarm methods. More recently, Liu \textit{et al.}~\cite{liu2025human} introduced a human-like strategy that considers AV–HV interactions, thereby improving performance in mixed-traffic settings.  

Despite these advances, existing research predominantly focuses on the host vehicle’s personality traits, while largely neglecting the intrinsic and interactive diverse intentions of surrounding human drivers. Yet these personalized cooperation intentions are critical for ensuring both safety and efficiency in lane change scenarios.  

To address this gap, we propose an intention-driven lane change framework that explicitly models the dynamic cooperation willingness of surrounding drivers in the target lane. By embedding both intrinsic and interactive personality-informed intentions into decision-making, the framework aims to achieve safer and more socially compliant AV maneuvers.

\section{Problem Statement}
To guarantee both the safety and efficiency of lane change maneuvers, it is essential to coordinate decisions based on the predicted personalized intentions of surrounding human drivers. Accordingly, we propose an intention-driven lane change framework that includes the recognition of human driving styles and intention prediction, as illustrated in Fig.~\ref{fig:framework}.

\subsection{Problem Formulation}

We consider a lane change scenario consisting of two adjacent lanes: the current lane occupied by the ego vehicle (AV, shown in red) and the target lane with surrounding human-driven vehicles (HVs, shown in blue and yellow). The ego vehicle’s decision is primarily influenced by two key HVs: the leading vehicle in the current lane and the target-rear (T-Rear) vehicle in the adjacent lane. Given the ego vehicle’s intention to merge, two candidate actions are defined: \emph{lane change} (LC) and \emph{lane keeping} (LK). Accordingly, the lane change decision-making problem is formulated as a binary classification task.  

To mimic human driving processes, we model the environment as a Markov decision process (MDP) with state $S$, action $A$, and reward $R$.  
The \textbf{state space} $S$ comprises both ego-vehicle states (velocity $v_E$, acceleration $a_E$) and surrounding-vehicle observations, including the velocities of the leading and target-rear vehicles $(v_f, v_{tr})$ and their relative distances to the ego vehicle $(d_{Ef}, d_{Etr})$. These features are grouped into \emph{inner features} (ego-centric) and \emph{inter features} (interaction-related).  
The \textbf{action space} $A = \{\textit{LC}, \textit{LK}\}$ corresponds to either initiating a lane change or maintaining the current lane.  
The \textbf{reward} $R$ represents the cumulative return defined by efficiency, safety, and comfort, and will be formally specified in Section~\ref{sec:methodology}.

The T-Rear vehicle plays a decisive role in determining whether the ego vehicle can safely execute a lane change. Its driving style, therefore, becomes a central factor in our framework, motivating the integration of human driver heterogeneity into both decision-making and motion-planning.  

For clarity, the lane change process can be divided into three phases:  
1) \textbf{Observation period}: the ego vehicle monitors surrounding states and infers the driving style of the T-Rear vehicle;  
2) \textbf{Decision period}: the ego vehicle selects between LC and LK based on observed features and cooperation intention;  
3) \textbf{Execution period}: once LC is chosen, the maneuver is carried out from initiation to completion.

\subsection{Dataset}
This study employs training and evaluation on the Next Generation SIMulation (NGSIM) I-80 dataset~\cite{halkia2006interstate} and cross-domain evaluation on highD dataset~\cite{krajewski2018highd}. The NGSIM dataset was collected on a section of the Interstate 80 freeway in California using multiple synchronized video cameras installed on nearby high-rise buildings. The recorded videos were subsequently processed to extract vehicle trajectories at a sampling rate of 10 Hz. The dataset provides detailed microscopic information such as vehicle position, velocity, and lane assignment, offering a naturalistic and high-resolution representation of mixed-traffic conditions. Owing to its comprehensiveness and accuracy, the NGSIM I-80 dataset has become one of the most widely adopted benchmarks for studying lane change behaviors and human–vehicle interactions. The highD dataset, released by RWTH Aachen University, captures naturalistic highway traffic from an aerial perspective using unmanned aerial vehicles. It contains trajectories of over 110,000 vehicles recorded at 25~Hz across six highway locations, covering 11.5 hours of driving and approximately 45,000~km in total. The dataset includes around 5,600 complete lane change maneuvers with a positioning error below 10~cm.

To obtain reliable lane change episodes, we preprocess the raw dataset in NGSIM I-80 with the following filtering criteria:  
\begin{itemize}
    \item Only small passenger cars traveling within lanes 1–4 are retained.  
    \item Each vehicle must exhibit exactly one lane change event within the monitored segment.  
    \item The longitudinal span of the maneuver is constrained between 300 ft and 1,900 ft.  
    \item A minimum duration of 10 seconds is preserved before and after the lane change event.  
    \item Trajectories of the ego vehicle and its adjacent front and rear vehicles are extracted for both pre- and post-maneuver phases.  
\end{itemize}

After filtering, a rolling median filter with a 5~s window and a Savitzky–Golay filter are sequentially applied to smooth the trajectories and reduce noise. Following these preprocessing steps, a total of 307 valid lane change episodes are extracted, each containing detailed trajectories of the ego vehicle and surrounding vehicles. Finally, to construct training samples, all trajectories are segmented into 2~s sliding windows, yielding a dataset of 5,295 samples in total: 3,666 LK samples and 1,629 LC samples.

Lane change start and end points are extracted during data preprocessing based on the lateral motion of each vehicle. Specifically, we compute the lateral velocity of the trajectory and identify the lane change interval by bidirectionally searching from the timestep with peak lateral speed until it falls below a predefined threshold $\delta = 0.2$. To reduce false detections caused by noise or minor lateral adjustments, we additionally enforce a lane-line continuity constraint, requiring consistent lane markings throughout the detected interval. The resulting start and end points define the lane change duration used in subsequent analysis.

\section{Methodology}
\label{sec:methodology}
This section presents our intention-driven decision-making and motion-planning framework for autonomous lane changes. The framework consists of four modules: (i) driving style clustering and recognition, (ii) intention prediction via cooperation scoring, (iii) decision-making with policy and reward networks, and (iv) motion-planning that generates executable trajectories. The overall workflow is shown in Fig.~\ref{fig:framework}.

\subsection{Driving Style Clustering and Recognition Module}

To enable adaptive decision-making, the driving style of surrounding vehicles is recognized from their historical motion patterns. We adopt a two-stage procedure: (i) unsupervised clustering to establish representative driving styles, and (ii) supervised recognition for online classification.

\begin{table}[!t]
\caption{Features selected for driving style clustering}
\label{ds_feature}
\renewcommand{\arraystretch}{1.3}
\centering
\small
\begin{tabular}{cc}
\toprule[1.2pt]
Symbol & Feature Description \\
\midrule
$\bar{v}$ & Average velocity \\
$\bar{a}$ & Average acceleration \\
$v_{std}$ & Std. deviation of velocity \\
$a_{std}$ & Std. deviation of acceleration \\
$v_{max}$ & Maximum velocity \\
$a_{max}$ & Maximum acceleration \\
\bottomrule[1.2pt]
\end{tabular}
\end{table}

\paragraph{Feature extraction and clustering}  
From each vehicle trajectory, six statistical features are computed, including mean and maximum velocity/acceleration and their standard deviations (see Table~\ref{ds_feature}). To alleviate dimensionality issues, features are transformed via principal component analysis (PCA). K-means clustering is then applied to group drivers into three canonical styles (aggressive, normal, conservative)~\cite{liu2022inverse, hao2025styledrivedrivingstyleawarebenchmarking}, by minimizing the intra-cluster squared Euclidean distance. This unsupervised step provides discrete style labels without requiring manual annotation.

\paragraph{Recognition model}  
Once clustering labels are obtained, a supervised classifier is trained to predict the style of new trajectories in real time. The network takes the same six features as input and consists of two hidden layers (64 units each, ReLU activation), followed by a softmax output over three style categories. The trained recognizer $f_{\text{style}}$ outputs probabilities $p(\text{style}|\mathbf{s}_t)$, which are subsequently used in the intention-prediction module.

This procedure ensures that latent driving styles are consistently extracted from raw behavioral data and can be integrated into the cooperation-aware decision-making framework.

\subsection{Intention-prediction Module}
The intention-prediction module quantifies the cooperative tendency of surrounding drivers through two complementary scores: a \emph{learnable intrinsic score} capturing style-dependent cooperation and a \emph{dynamic interactive score} derived from inter-vehicle interactions. These scores are subsequently fused through a trainable gating mechanism.

\paragraph{Learnable Cooperation Score (LCS)}  
The \textbf{intrinsic cooperation score} encodes cooperation tendencies specific to the target vehicle’s internal driving style. It is computed by concatenating the inner features $\mathbf{x}$ (e.g., velocity, acceleration) with the style label $s$, and projecting them through a learnable encoder $f_{\text{intr}}$:
\begin{equation}
c^{\text{intr}}_t = f_{\text{intr}}([\mathbf{x};\,s];\,\theta_{\text{intr}}).
\end{equation}

\paragraph{Dynamic Cooperation Score (DCS)}  
The \textbf{interactive cooperation score} reflects how a driver responds to its surrounding context, and is obtained from interaction-level features $\mathbf{h}$ (e.g., relative distance, relative velocity):
\begin{equation}
c^{\text{inter}}_t = f_{\text{inter}}(\mathbf{h};\,\theta_{\text{inter}}).
\end{equation}
Unlike LCS, DCS adapts to momentary traffic interactions, complementing the short-term style-driven characterization.
The parameters $\theta_{\text{intr}}$ and $\theta_{\text{inter}}$ are trainable and receive gradients both from the cooperation regularization loss and from the downstream decision-making objective, ensuring that the learned score reflects intrinsic cooperation characteristics consistent with safe and efficient driving. Note that both $c^{\text{intr}}_t$ and $c^{\text{inter}}_t$ are constrained to the range (0, 1) via a sigmoid activation.

The final cooperation score is produced through a learnable gate that dynamically balances the contributions of LCS and DCS:
\begin{equation}
c^{\text{final}}_t = \alpha_t \cdot c^{\text{intr}}_t + (1-\alpha_t)\cdot c^{\text{inter}}_t,
\end{equation}
with gating coefficient
\begin{equation}
\alpha_t = \sigma\!\big(\text{MLP}([\mathbf{x},\,\mathbf{h}])\big).
\end{equation}
The gate is fully differentiable and updated by gradient backpropagation, enabling context-dependent modulation of cooperation.

\begin{algorithm}[!t]
\DontPrintSemicolon
\small
\caption{Training the Intention-driven Lane Change Decision-making Model}
\label{alg:train_coop_clear}
\KwIn{Dataset $\mathcal{D}=\{(\mathbf{x},\mathbf{h},s,a)\}$ with inner features $\mathbf{x}$, inter features $\mathbf{h}$, style label $s$, and binary action label $a$.
Hyperparameters: learning rate $\eta$, batch size $B$, epochs $E$, reward temperature $\beta$, $\lambda_{L2}$ , $\lambda_s$, loss function weights $\lambda_1$ and $\lambda_2$}.
\KwOut{Trained $f_{\text{intr}}$, $f_{\text{inter}}$, $f_\pi$, $f_r$.}

\BlankLine
Initialize $f_{\text{intr}}, f_{\text{inter}}, f_\pi, f_r$ and Adam optimizers with learning rate $\eta$.\;

\For{$epoch \gets 1$ \KwTo $E$}{
  \ForEach{mini-batch $\{(\mathbf{x}_i,\mathbf{h}_i,s_i,a_i)\}_{i=1}^{B}$}{
    
    Compute LCS: $c^{\text{intr}}_i \gets f_{\text{intr}}([\mathbf{x}_i; s_i]; \theta_\text{intr})$,\;
    Compute DCS: $c^{\text{inter}}_i \gets f_{\text{inter}}(\mathbf{h}_i; \theta_\text{inter})$,\;
    $\alpha_i \gets \sigma(\text{MLP}([\mathbf{x}_i,\mathbf{h}_i]))$,\;
    $c^{\text{final}}_i \gets \alpha_i c^{\text{intr}}_i + (1-\alpha_i)c^{\text{inter}}_i$.\;
    
    Predict decision: $z_i \gets f_\pi(\mathbf{x}_i,\mathbf{h}_i,c^{\text{final}}_i)$,\;
    $\ell^{(i)}_{\mathrm{BCE}} \gets -[a_i\log\sigma(z_i)+(1-a_i)\log(1-\sigma(z_i))]$.\;
    
    Predict rewards: $r^{(1)}_i \gets f_r(\mathbf{x}_i,\mathbf{h}_i,c^{\text{final}}_i,1)$; $r^{(0)}_i \gets f_r(\mathbf{x}_i,\mathbf{h}_i,c^{\text{final}}_i,0)$,\;
    $w_i \gets \frac{\sigma(\beta(r^{(1)}_i-r^{(0)}_i))}{\mathbb{E}[\sigma(\beta(r^{(1)}_i-r^{(0)}_i))]}$.\;
    
    $\mathcal{L}_{\mathrm{BC}} \gets \frac{1}{B}\sum_i w_i\,\ell^{(i)}_{\mathrm{BCE}}$.\;
    
    $\mathcal{L}_{\mathrm{IRL}} \gets -\frac{1}{B}\sum_i \Big[r^{(a_i)}_i - \log(\exp(r^{(a_i)}_i)+\exp(r^{(1-a_i)}_i))\Big]$\\
    $\quad +\, \lambda_{L2} \frac{1}{B}\sum_i\Big[(r^{(a_i)}_i)^2+(r^{(1-a_i)}_i)^2\Big]$\\
    $\quad +\, \lambda_s \frac{1}{B}\sum_i \big|r^{(a_i)}_i-r^{(1-a_i)}_i\big|$.\;
    
    $\mathcal{L}_{\mathrm{coop}} \gets \frac{1}{B}\sum_i (c^{\text{final}}_i-0.5)^2$.\;
    
    $\mathcal{L} \gets \lambda_1\mathcal{L}_{\mathrm{BC}}+\lambda_2\mathcal{L}_{\mathrm{IRL}}+\mathcal{L}_{\mathrm{coop}}$.\;
    
    Update $\{\theta_{\text{intr}},\theta_{\text{inter}},\theta_\pi,\theta_r\}$ with Adam using $\nabla \mathcal{L}$.\;
  }
}
\Return $f_{\text{intr}}, f_{\text{inter}}, f_\pi, f_r$.
\end{algorithm}

\subsection{Decision-making Module}

The decision-making module integrates the fused cooperation score $c^{\text{final}}_t$ with both inner and interaction features $\{\mathbf{x}_t,\mathbf{h}_t\}$ to jointly learn a policy and a reward function. The whole algorithm can be found in Algorithm~\ref{alg:train_coop_clear}

\begin{itemize}
    \item \textbf{Policy Net} $f_\pi$: an LSTM-based classifier that outputs the probability $\hat{a}_t$ of executing a lane change versus keeping lane, trained with a reward-weighted behavior cloning loss.  
    \item \textbf{Reward Net} $f_r$: an auxiliary network optimized via inverse reinforcement learning, which estimates latent rewards $r^{(a)}_t$ for each candidate action $a$. These rewards are used both to shape the policy’s weighted imitation and to enforce discriminability between cooperative and non-cooperative behaviors.  
\end{itemize}

The overall training objective combines three components:

\begin{equation}
L_{total} = \lambda_1 \cdot L_{BC} + \lambda_2 \cdot L_{IRL} + L_\mathrm{coop},
\end{equation}
where $\lambda_1$ and $\lambda_2$ are the weights of BC and IRL loss. The weights of different loss components and regularization terms are summarized in Table~\ref{tab:training_and_loss} and are kept fixed across all experiments. Other terms are defined as follows.

\paragraph{Reward-weighted Behavior Cloning (Explored Design)}
As an alternative design, we investigated a tightly-coupled reward-weighted behavior cloning formulation,
in which the learned reward function directly modulates the contribution of each training sample.
Given a training instance $(\mathbf{x}_i,\mathbf{h}_i,s_i,a_i)$, where $a$ is the binary action label, the policy predicts a logit $z_i = f_\pi(\mathbf{x}_i,\mathbf{h}_i,c_i^{\text{final}})$,
and the per-sample BCE loss is defined as
\begin{equation}
\ell^{(i)}_{\mathrm{BCE}} = -\Big[a_i \log\sigma(z_i) + (1-a_i)\log(1-\sigma(z_i))\Big].
\end{equation}
The loss is reweighted by a reward-dependent coefficient
\begin{equation}
w_i = \frac{\sigma\big(\beta(r^{(1)}_i - r^{(0)}_i)\big)}
{\tfrac{1}{B}\sum_{j=1}^{B}\sigma\big(\beta(r^{(1)}_j - r^{(0)}_j)\big)},
\end{equation}
where $\beta$ denotes the reward temperature and
$r^{(1)}_i, r^{(0)}_i$ correspond to the reward estimates for lane change and lane keeping, respectively.
The objective is
\begin{equation}
L_{\mathrm{BC}} = \frac{1}{B}\sum_{i=1}^B w_i\,\ell^{(i)}_{\mathrm{BCE}}.
\end{equation}

\paragraph{Inverse Reinforcement Learning (IRL) loss}
For each instance, the reward net evaluates both the chosen action $a_i$ and the alternative action $1-a_i$:
\begin{equation}
\begin{aligned}
    r^{(a_i)}_i &= f_r(\mathbf{x}_i,\mathbf{h}_i,c^{\text{final}}_i,a_i),
    \quad \\ r^{(1-a_i)}_i &= f_r(\mathbf{x}_i,\mathbf{h}_i,c^{\text{final}}_i,1-a_i).\end{aligned}
\end{equation}
The IRL objective enforces preference consistency, with additional regularization:
\begin{align}
L_{\mathrm{IRL}} \;=\;& -\frac{1}{B}\sum_{i=1}^B \Big[r^{(a_i)}_i - \log\big(\exp(r^{(a_i)}_i)+\exp(r^{(1-a_i)}_i)\big)\Big] \nonumber \\
&+ \lambda_{L2} \frac{1}{B}\sum_{i=1}^B \Big[(r^{(a_i)}_i)^2+(r^{(1-a_i)}_i)^2\Big] \nonumber \\
&+ \lambda_s \frac{1}{B}\sum_{i=1}^B \big|r^{(a_i)}_i-r^{(1-a_i)}_i\big|.
\end{align}

\paragraph{Cooperation regularization loss}
To avoid degenerate cooperation behaviors and stabilize training, we introduce a soft regularization term on the final cooperation score. Specifically, we penalize large deviations from a neutral cooperation level:
\begin{equation}
L_{\mathrm{coop}} = \frac{1}{B}\sum_{i=1}^B \big(c^{\text{final}}_i - 0.5\big)^2.
\end{equation}

This regularization does not enforce a fixed cooperation value. Instead, the target value $0.5$ serves as a neutral prior that discourages the cooperation score from collapsing to extreme values (0 or 1), which we empirically observed to reduce interaction diversity and degrade robustness in mixed-traffic scenarios. By acting as a soft constraint, $L_{\mathrm{coop}}$ preserves the model’s ability to adapt cooperation levels based on contextual interaction cues, while preventing trivially always-cooperative or always-noncooperative solutions.

The overall objective remains fully differentiable with respect to all parameters $\{\theta_{\text{intr}}, \theta_{\text{inter}}, \theta_\pi, \theta_r\}$, and the model is optimized end-to-end using the Adam optimizer.

\subsection{Motion-planning Module}

\paragraph{IRL-based target vehicle trajectory prediction}
To predict the motion of surrounding vehicles, we adopt maximum-entropy inverse reinforcement learning (Max-Ent IRL)~\cite{ziebart2008maximum}. Unlike behavior cloning, which directly maps states to actions, IRL infers the latent reward function $R(s,a)$ that best explains expert demonstrations. Under the Max-Ent principle, the probability of a trajectory $\zeta$ is
\begin{equation}
\begin{aligned}
    p(\zeta \mid \omega) &= \frac{1}{Z(\omega)}\exp(\sum_tr(s_t)) \\
    &\approx \frac{\exp\big(\sum_t \omega^T \textbf{f}_\zeta\big)}{\sum_{\zeta'} \exp\big(\sum_t \omega^T \textbf{f}_{\zeta'}\big)},\end{aligned}
\end{equation}
where $\omega$ is the reward weight vector, $Z(\omega) = \sum_{\zeta'} exp(\sum_tr)$ is the partition function, $r$ is the cumulative reward, $\textbf{f}_\zeta$ is the feature vector of $\zeta$, and $\zeta'$ represents a candidate trajectory. Parameters in $\omega$ are identified and learned by maximizing the log-likelihood of expert trajectories:
\begin{equation}
\max_\omega L(\omega) = \sum_{\zeta \in D}\log p(\zeta \mid \omega).
\end{equation}

\paragraph{Reward function design}

Following established formulations~\cite{qiu2024driving}, the reward function is modeled as a linear combination of interpretable driving features:
\begin{equation}
R(s,a) = \omega^T \Phi(s,a),
\end{equation}
where the feature vector $\Phi(s,a)$ encodes three fundamental aspects of human driving behavior: {Efficiency}, {Safety}, and {Comfort}.

The \textbf{Efficiency} feature, $f_{\text{eff}}(t)$, reflects the vehicle’s velocity magnitude, defined as:
\begin{equation}
f_{\text{eff}}(t) = |v_{\text{ego}}(t)|.
\end{equation}

The \textbf{Safety} feature, $f_{\text{safety}}(t)$, is designed to penalize short headways, ensuring a safe following distance, formulated as:
\begin{equation}
f_{\text{safety}}(t) = \exp\!\left(-\frac{x_{lv}(t)-x_{ego}(t)}{v_{ego}(t)+\epsilon}\right),
\end{equation}
where $x_{lv}$ and $x_{ego}$ denote the positions of the leading and ego vehicles, respectively, and $\epsilon$ is a small constant introduced to prevent division by zero.

The \textbf{Comfort} feature, $f_{\text{comfort}}(t)$, discourages abrupt maneuvers by incorporating the jerk (the derivative of acceleration, $\dot{a}_{ego}(t)$) into its definition:
\begin{equation}
f_{\text{comfort}}(t) = 1-\exp\!\big(-|\dot{a}_{ego}(t)|\big),
\end{equation}
where $\dot{a}_{ego}(t)$ is the jerk of the ego vehicle at time $t$.

Thus, the reward is accumulated as
\begin{equation}
R(s,a) = \omega_{\text{eff}} f_{\text{eff}} + \omega_{\text{safety}} f_{\text{safety}} + \omega_{\text{comfort}} f_{\text{comfort}},
\end{equation}
where the weights $\{\omega_{\text{eff}},\omega_{\text{safety}},\omega_{\text{comfort}}\}$ are learned via Max-Ent IRL to balance efficiency, safety, and comfort in human-like trajectory prediction.

\paragraph{MPC-based motion-planning and control}
With the lane change command of the proposed decision-making model, the AV is able to generate a lane change path based on the predicted trajectory of the target vehicle. Then MPC is used to control the AV drive from the current lane to the target lane considering the constraint of collision avoidance with the surrounding vehicles.
Given the lane change command and the predicted trajectory of the following vehicle in the target lane, we generate a smooth lane change trajectory based on the Sigmoid function.
To sufficiently utilize the predicted trajectory of the target vehicle, we generate the reference path of the AV using the Sigmoid function inspired by~\cite{ammour2020trajectory}.
With the input of the predicted trajectory of the target vehicle, and the current and target state of the ego vehicle, the expression of the sigmoid function-based trajectory is elaborated as follows.
The sigmoid function-based trajectory incorporates the predicted trajectory of the target vehicle along with the current and target states of the ego vehicle.
With the expected longitudinal distance $d_\text{long}$, lateral distance $d_\text{lat}$ during the lane change process, and the value of a dimensionless curvature-related variable $\tau$, the reference path is generated by:

\begin{equation}
\begin{aligned}
\tilde{x}(x) & =\hat{x}-\frac{d_{\text {long }}}{2}, \\
y(\tilde{x}) & =\frac{d_{\text {lat }}}{1+e^{-\tau(\tilde{x}+b)}}+y_0,
\end{aligned}\label{eq:laneChangeCurve}
\end{equation}
where $\tilde{x}(x)$ is an intermediate variable, given $\hat{x}\in [0, d_{\text{long}}]$. With $x(\hat{x})=\hat{x}+x_0$, we can generate a reference path for the AV to follow during the lane change process. Note that $\tau>0$ in (\ref{eq:laneChangeCurve}) serves as the shape parameter that regulates the steepness of the reference path curve. It dictates the rapidity of the transition between the lower asymptote, $y_0$, and the upper boundary $d_{\text{lat}}$.

The expected lateral distance $d_\text{lat}$ is predefined as the lane width, which is $3.6576$\,m by default. In contrast, the expected longitudinal distance is calculated by $d_\text{long}=\Bar{v}_\text{T-Rear}N\Delta T$, where $\Bar{v}_\text{T-Rear}$ is the average velocity of the target vehicle based on the predicted trajectory using a specific method, $N$ is the number of simulation steps, $\Delta T=0.1$\,s is the simulation time interval. Concretely, the average velocity of the target vehicle can be calculated by differentiating its predicted positions generated by the trajectory prediction methods. Therefore, the trajectory prediction of the target vehicle is vital for the lane change behavior of the AV.

The generated reference trajectory is then integrated into the MPC controller for trajectory tracking and safety guarantee.
We use a bicycle model to simulate the dynamics of the ego vehicle. Afterward, MPC is used to track the reference trajectory for lane change maneuvers. Next, following the recipe proposed in~\cite{liu2023integrated}, each line of the road and the surrounding vehicles are modeled using potential functions integrated into the MPC-based trajectory.

\begin{table}[t]
\centering

\small
\caption{Training Hyperparameters and Loss-Related Coefficients}
\label{tab:training_and_loss}
\begin{tabular}{lcc}
\toprule
Category & Symbol & Value \\
\midrule
\multicolumn{3}{l}{\textit{Training Hyperparameters}} \\
\midrule
Optimizer            & --            & Adam \\
Learning Rate        & $\eta$         & 0.002 \\
Weight Decay         & --             & $5 \times 10^{-4}$ \\
Batch Size           & $B$            & 64 \\
Epochs               & $E$            & 100 \\
\midrule
\multicolumn{3}{l}{\textit{Loss Function Weights and Regularization}} \\
\midrule
BC Loss Weight                    & $\lambda_1$     & 1.0 \\
IRL Loss Weight                   & $\lambda_2$     & 0.25 \\
IRL L2 Regularization Coefficient & $\lambda_{L2}$  & 0.001 \\
IRL Smoothness Regularization     & $\lambda_s$     & 0.01 \\
\bottomrule
\end{tabular}
\end{table}

\begin{figure}[!t]
  \centering
  \includegraphics[width=\linewidth]{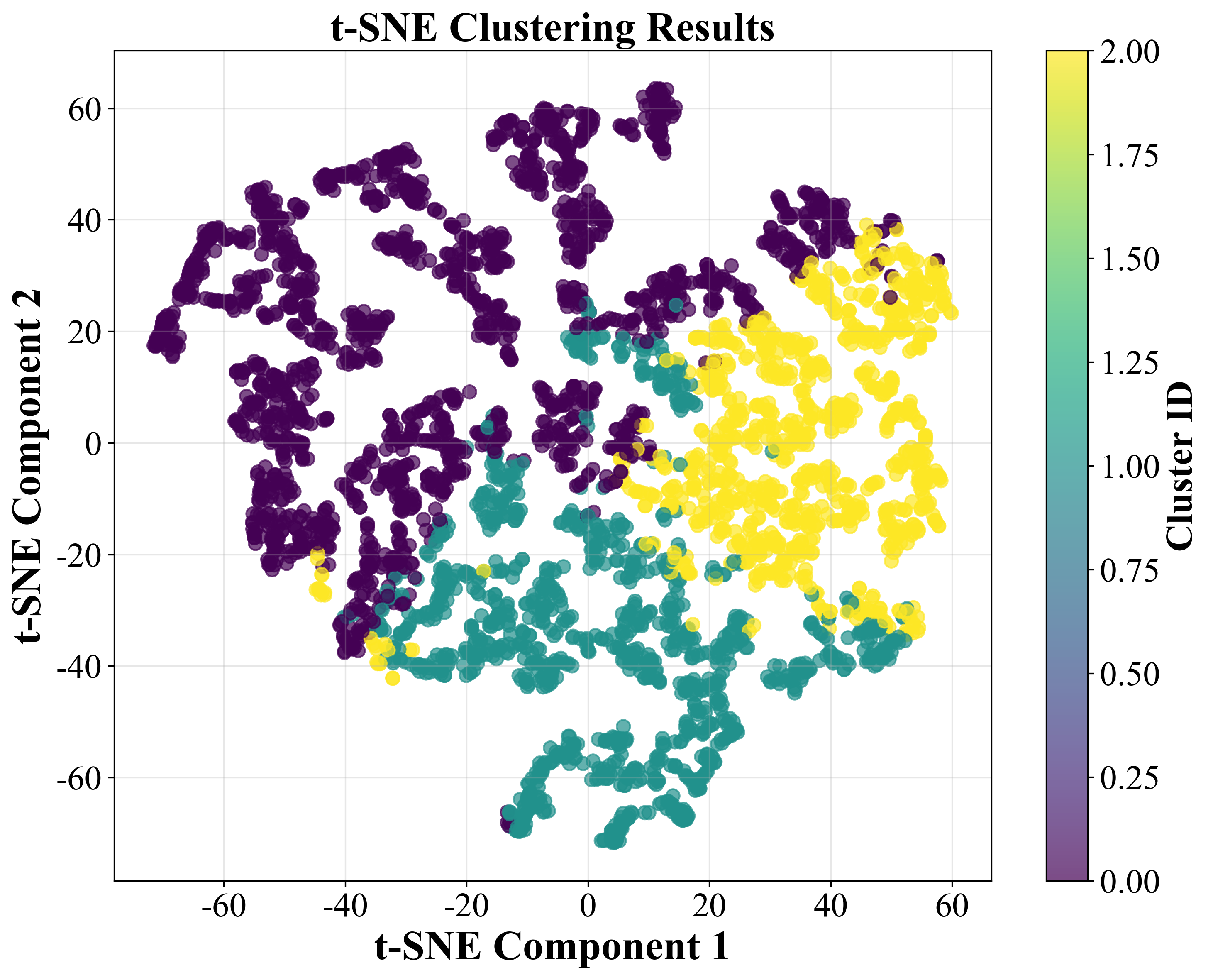}
    \label{fig:tsne}
  \caption{Analysis of driving-style representation: clustering results show distinct style separation.}
  \label{fig:clustering_analysis}
\end{figure}
\section{Experiments and Results}
\subsection{Implementation Details}
All models are trained using the Adam optimizer with a learning rate of 0.002, a weight decay of $5 \times 10^{-4}$, a batch size of 64, and a maximum of 100 training epochs (illustrated in Table~\ref{tab:training_and_loss}). For sequence modeling, the hidden dimension of the LSTM layers was set to 64. After preprocessing, a total of 5,295 samples were retained. The dataset was divided into training and validation subsets with an 80/20 split, where stratified sampling was employed to maintain class balance between LK and LC. Each sample consisted of 10-dimensional state features (containing inner- and inter-features) over a sequence length of 20 frames.

For the reward-weighted BC method, we conduct a sensitivity study on the reward temperature $\beta$ to evaluate the effect of reward-based sample reweighting.
Results show that setting $\beta=0$, which reduces the formulation to standard behavior cloning, achieves the best performance, while increasing $\beta$ leads to a degradation.
This indicates that directly using reward estimates for sample-level importance weighting introduces training instability rather than improvement. The complete sensitivity results are reported in the Appendix. These findings motivate our final design choice, in which behavior cloning and IRL objectives are optimized in a loosely-coupled manner.

\begin{table*}[!t]
\centering
\small

\caption{Comparison of different methods on lane keeping (LK) and lane change (LC) classification.}
\setlength{\tabcolsep}{4.5pt}
\begin{tabular}{lcccccccccc}
\toprule
\multirow{2}{*}{Methods} 
& \multicolumn{4}{c}{overall metrics} 
& \multicolumn{3}{c}{LK} 
& \multicolumn{3}{c}{LC} \\
\cmidrule(lr){2-5} \cmidrule(lr){6-8} \cmidrule(lr){9-11}
& accuracy & precision & recall & F1-score 
& precision & recall & F1-score 
& precision & recall & F1-score \\
\midrule
IDM+MOBIL    & 0.6185	 &0.6394&	0.6185&	0.6269&	0.7459&	0.6808	&0.7118	&0.4000	&0.4785	&0.4358 \\
H-LSTM & 0.9112&	0.9159&	0.9112&	0.9124	&0.9584	&0.9113	&0.9343	&0.8204	&0.9110	&0.8634 \\
BC      & 0.8461	&0.8502	&0.8461	&0.8477	&0.9034	&0.8728	&0.8878	&0.7275	&0.7844	&0.7549 \\
VWC  & 0.9207	&0.9219	&0.9207	&0.9211	&0.9513	&0.9332	&0.9421	&0.8559	&0.8926	&0.8739 \\ 
GNN & 0.9594  &0.9597   &0.9594  &0.9595 &\underline{0.9752}    &0.9659  &0.9705 &0.9249   &\underline{0.9448}  &0.9347\\
RNN  &0.9547  &0.9554  &0.9547  &0.9549 &0.9750 &0.9591  &0.9670 &0.9112 &\underline{0.9448}  &0.9277 \\
Transformer &\underline{0.9651}  &\underline{0.9650}  &\underline{0.9651}  &\underline{0.9650} &0.9728  &\textbf{0.9768}  &\underline{0.9748} &\textbf{0.9474}  &0.9387 &\underline{0.9430} \\
\rowcolor{gray!15}
Ours         & \textbf{0.9698}  &\textbf{0.9700}  &\textbf{0.9698} &\textbf{0.9699} &\textbf{0.9821}  &\underline{0.9741}  &\textbf{0.9781} &\underline{0.9428}  &\textbf{0.9601} &\textbf{0.9514} \\ 
\bottomrule
\end{tabular}
\label{tab:comparison}
\end{table*}

\begin{figure*}[!t]
  \centering
  
  \subfloat[Training and validation loss curves.]{
    \includegraphics[width=0.5\textwidth]{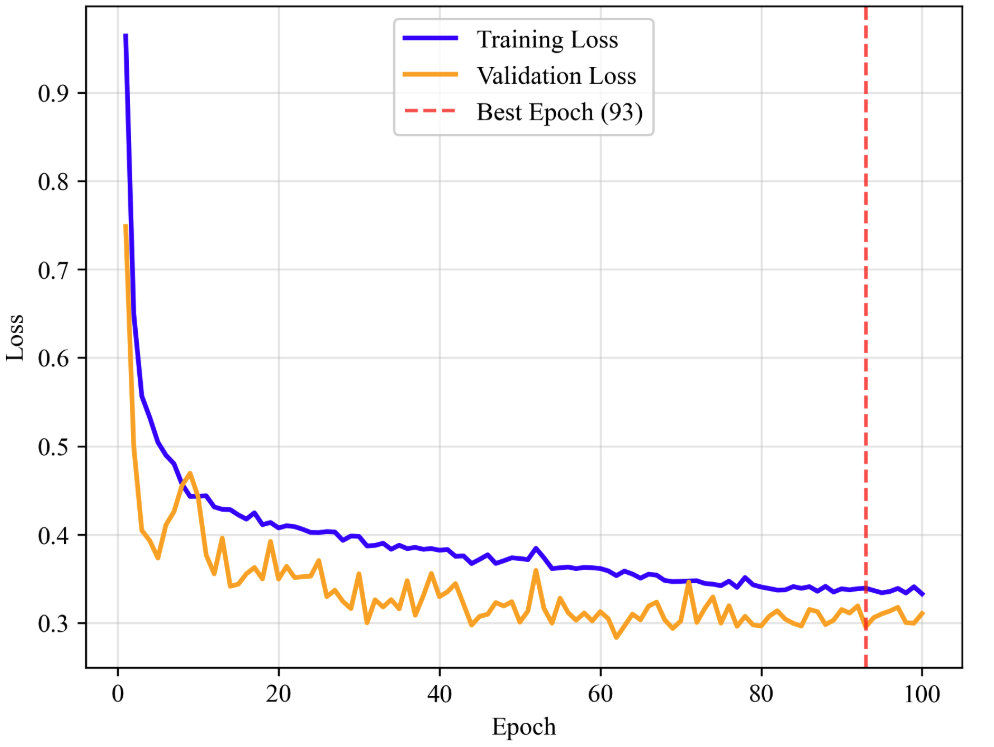}
    \label{fig:loss}
  }
  \hfill
  \subfloat[Confusion matrix for lane change classification results.]{
    \includegraphics[width=0.42\textwidth]{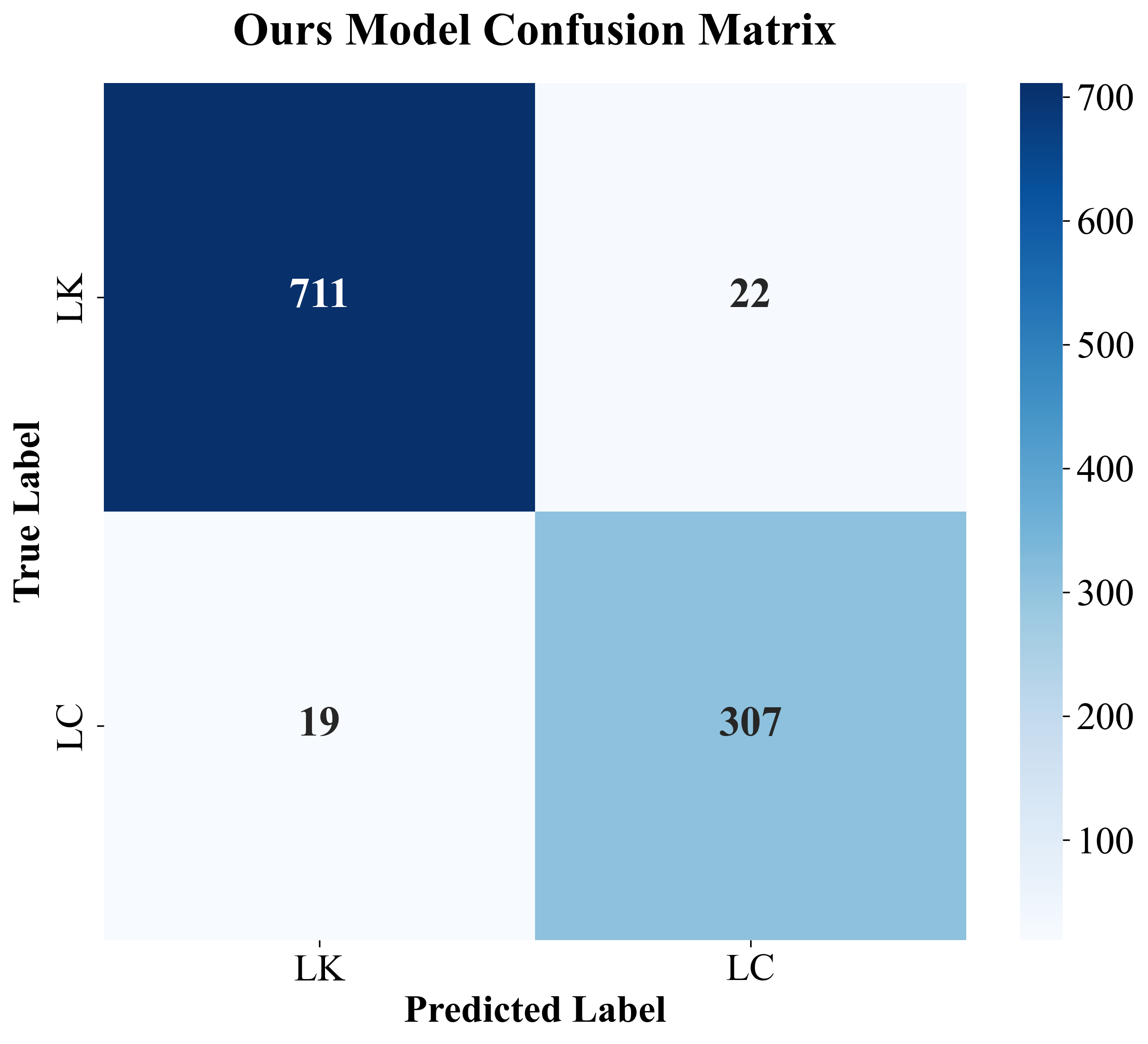}
    \label{fig:cm}
  }
  \caption{Performance evaluation of the proposed intention-driven lane change decision-making model, including training dynamics, validation metrics, and 
 outcomes.}
  \label{fig:decision_results}
\end{figure*}

\begin{table}[t]
\centering

\small
\caption{Learning-based Methods Inference Time Comparison}
\label{tab:inference_time}
\begin{tabular}{lc}
\toprule
Method & Inference Time (ms)  \\
\midrule
H-LSTM         & 19.53 $\pm$ 0.48  \\
BC             & 18.42 $\pm$ 0.36  \\
VWC            & 30.94 $\pm$ 0.57 \\
GNN            & 30.39 $\pm$ 0.59 \\
RNN         & 26.67 $\pm$ 0.66 \\
Transformer & 40.93 $\pm$ 0.70 \\
\textbf{Ours}  & \textbf{13.93 $\pm$ 0.28}\\
\bottomrule
\end{tabular}

\vspace{2pt}
\footnotesize{\textit{Note:} Inference time is measured over 100 runs with batch size 100.}
\end{table}

\begin{table}[t]
\centering
\small

\caption{Learning-based Methods Cross-Domain Evaluation on highD dataset.}
\label{tab:cross_domain_highd_zero_shot}
\begin{tabular}{lcccc}
\toprule
Methods & Accuracy & Precision & Recall & F1-score \\
\midrule
H-LSTM         & 0.8044 & 0.7176 & 0.8044 & 0.7542 \\
BC             & 0.7820 & 0.7070 & 0.7820 & 0.7410 \\
VWC            & \textbf{0.8290} & 0.7177 & \textbf{0.8290} & 0.7620 \\
GNN            & 0.8217 & 0.7108 & 0.8217 & 0.7583 \\
RNN         & 0.8060 & 0.7205 & 0.8060 & 0.7559 \\
Transformer & 0.7826 & 0.7072 & 0.7826 & 0.7413 \\
\textbf{Ours}  & \underline{0.8228} & \textbf{0.7536} & \underline{0.8228} & \textbf{0.7733} \\
\bottomrule
\end{tabular}
\end{table}

\subsection{Main Results}
The clustering results are shown in Fig.~\ref{fig:clustering_analysis}, the t-SNE manifests that the three groups are mostly separated well. 
To validate the impact of integrating human driving styles into the decision-making process, we compare the proposed model with representative rule-based and imitation-learning baselines, as well as advanced sequence modeling and interaction-aware architectures, including RNN-, GNN-, and Transformer-based methods.:

\begin{itemize}
    \item \textbf{IDM+MOBIL:} A rule-based baseline that integrates the Intelligent Driver Model (IDM) for longitudinal control~\cite{kesting2010enhanced} with the MOBIL framework for lateral maneuvers~\cite{kesting2007general}. Lane change decisions are triggered by predefined safety constraints and acceleration incentives, aiming to balance efficiency and safety.
    
    \item \textbf{H-LSTM:} A learning-based approach employing a hierarchical long short-term memory (H-LSTM) network~\cite{liao2022online} to capture temporal dependencies in multi-vehicle interactions and enable online lane change prediction.
    
    \item \textbf{BC:} A supervised imitation-learning baseline~\cite{wang2023high}, where the policy directly maps state features to expert actions via behavior cloning, without explicit reward modeling.
    \item \textbf{VWC:} A learning-based baseline employing a variable-weight cooperation (VWC) model~\cite{wang2024variable}, which adaptively assigns dynamic weights to lane change conditions and driving demands across different phases of interaction.
    \item \textbf{GNN:} A graph neural network(GNN)-based baseline that models multi-vehicle interactions through message passing over dynamically constructed traffic graphs. 
    \item \textbf{RNN:} A recurrent neural network (RNN) baseline that predicts lane change decisions from temporal sequences of observed traffic states.     
    \item \textbf{Transformer:} A Transformer-based baseline~\cite{gao2023dual} that employs self-attention to aggregate historical trajectories and interaction context for lane change decision-making.
    
\end{itemize}

Evaluation metrics for lane change intention prediction include precision, recall, and the F1 score, defined as follows~\cite{frossard2019deepsignals}:
\begin{linenomath}
    \begin{align}
        \text{precision} &= \frac{TP}{TP+FP}, \\
        \text{recall} &= \frac{TP}{TP+FN}, \\
        \text{F1} &= 2 \times \frac{\text{recall} \times \text{precision}}{\text{recall} + \text{precision}},
    \end{align}
\end{linenomath}
where $TP$ denotes true positives, samples correctly predicted as $LC$; $FP$ denotes false positives, $LK$ samples incorrectly predicted as $LC$; $FN$ denotes false negatives, $LC$ samples incorrectly predicted as other decisions.

Table~\ref{tab:comparison} compares the performance of different baselines against our proposed method on LK and LC classification. The results clearly show that rule-based IDM+MOBIL performs the worst, with overall accuracy of only 0.6185, reflecting its limited adaptability to heterogeneous human driving patterns. Learning-based methods achieve significantly higher performance: H-LSTM reaches above 0.91 across all metrics by capturing temporal dependencies, while BC attains moderate performance (accuracy 0.8461) but is constrained by its reliance on direct imitation. The VWC model further improves classification accuracy to 0.9207 by adaptively weighting interactive features. More recent interaction-aware architectures demonstrate further gains. GNN and RNN models achieve overall accuracies of 0.9594 and 0.9547, respectively, indicating the effectiveness of explicitly modeling inter-vehicle dependencies and temporal dynamics. The Transformer-based model achieves 0.9651 accuracy, benefiting from its global attention mechanism for capturing long-range interactions. In contrast, our proposed intention-driven BC-IRL framework achieves the best overall results, with accuracy 0.9698 and F1-score 0.9699. Notably, it further improves lane change recognition performance, demonstrating the advantage of explicitly modeling heterogeneous cooperation through intrinsic and interaction-driven components.

Fig.~\ref{fig:decision_results} further illustrates the training dynamics and classification outcomes. Fig.~\ref{fig:decision_results}(a) shows that both training and validation losses decrease steadily, with no significant overfitting, indicating stable convergence. The confusion matrix in Fig.~\ref{fig:decision_results}(b) shows that the model achieves high true positive and true negative rates, with only a small number of misclassifications, confirming the effectiveness of the intention-driven decision-making framework in distinguishing LK and LC intentions.
As shown in Table~\ref{tab:inference_time}, our method exhibits the lowest inference time among all learning-based baselines.
Despite incorporating interaction modeling, it maintains a substantially lower latency than RNN-, GNN-, and Transformer-based approaches.

\begin{table*}[!t]
\centering

\small
\caption{Ablation study results on the proposed framework.}
\setlength{\tabcolsep}{4.2pt}
\begin{tabular}{ccccccccc}
\toprule
\multicolumn{2}{c}{Core Module} & 
\multicolumn{2}{c}{Coop. Module} & 
\multicolumn{1}{c}{Style-aware} &
\multicolumn{4}{c}{Metrics} \\
\cmidrule(lr){1-2} \cmidrule(lr){3-4} \cmidrule(lr){5-5} \cmidrule(lr){6-9}
BC & IRL & LCS & DCS & w/o Style & Accuracy & Precision & Recall & F1-score \\
\midrule
\checkmark &   &   &   &  & 0.9301 &  0.9328 & 0.9301 &  0.9308 \\
\checkmark & \checkmark &  &  &  &  0.9339 &  0.9339 &  0.9339&   0.9339 \\
\checkmark & \checkmark & \checkmark &   &  & 0.9282 & 0.9357 & 0.9282 & 0.9296 \\
\checkmark & \checkmark &   & \checkmark & & 0.9254 &0.9336&    0.9254 &0.9268 \\
\rowcolor[gray]{0.9}
\checkmark & \checkmark & \checkmark & \checkmark &   & \textbf{0.9509} & \textbf{0.9529} & \textbf{0.9509} & \textbf{0.9514} \\
\checkmark & \checkmark & \checkmark & \checkmark & \checkmark &
 0.9471 &0.9528 &   0.9471 &0.9480
\\
\bottomrule
\end{tabular}
\label{tab:ablation}
\end{table*}

\begin{figure*}[!t]
  \centering
  
  \subfloat[Composition of final cooperation score according to different driving styles.]{
    \includegraphics[width=0.52\linewidth]{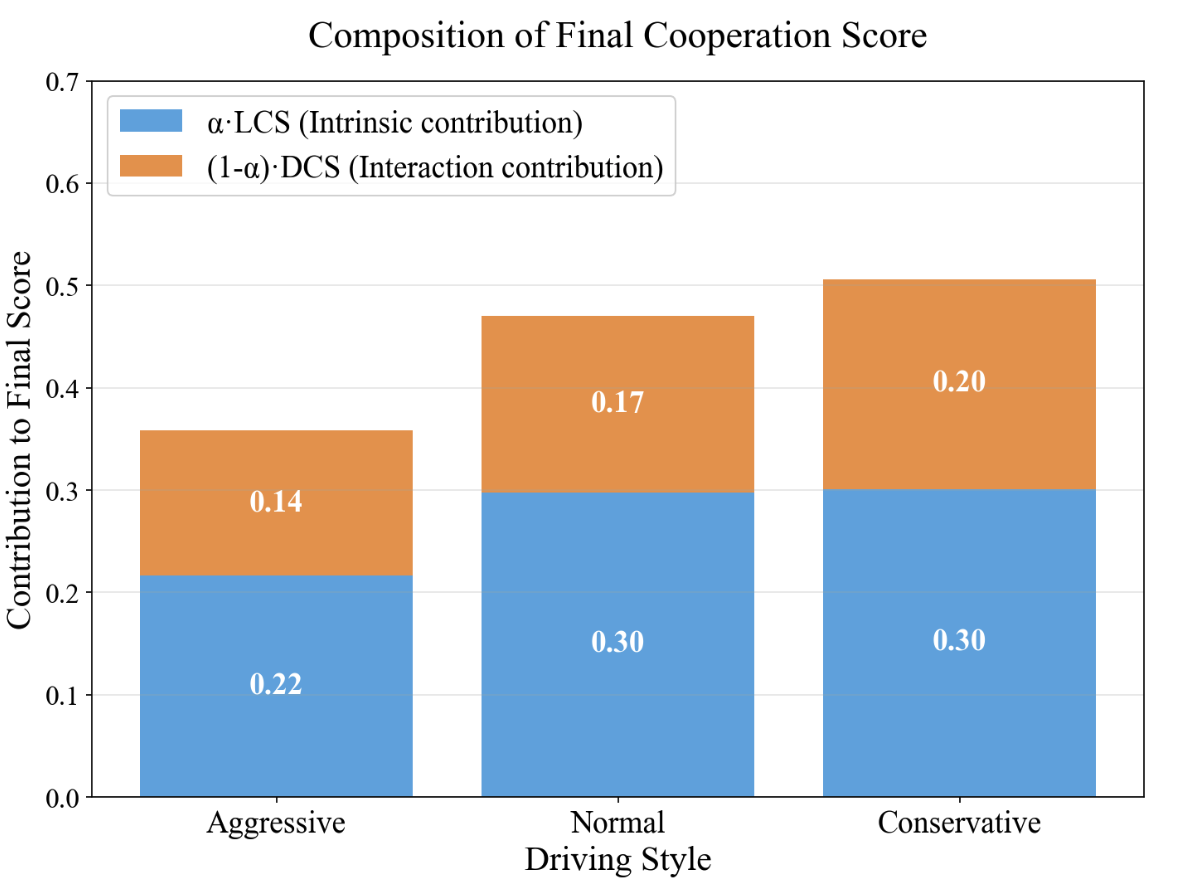}
    \label{fig:final_cop_analysis}
  }
  \hfill
  \subfloat[Cooperation profile by driving style.]{
    \includegraphics[width=0.42\linewidth]{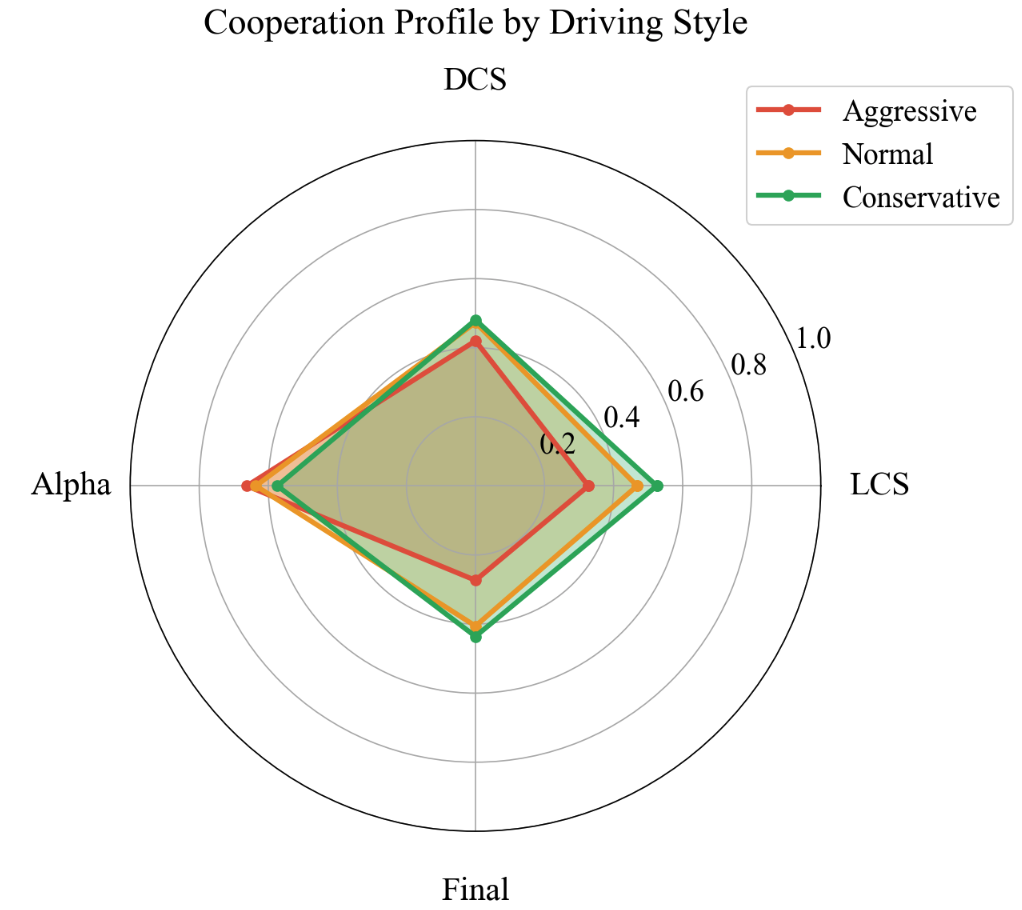}
    \label{fig:profiles}
  }
  \caption{Analysis of cooperation-aware parameters: (a) composition of final cooperation score including intrinsic and interaction cooperation scores (LCS/DCS); (b) cooperation profile analysis of different driving style.}
  \label{fig:coo_analysis}
\end{figure*}

To further evaluate the generalization ability of different learning-based methods, we conduct a cross-domain evaluation on the highD dataset using the same preprocessing and evaluation protocol, without any retraining or dataset-specific fine-tuning. The results are summarized in Table~\ref{tab:cross_domain_highd_zero_shot}.
Compared with in-domain performance on NGSIM, all methods exhibit a performance drop on highD due to differences in traffic density, road geometry, and data collection viewpoints, highlighting the challenge of cross-dataset lane change recognition. Among the baselines, VWC achieves the highest accuracy (0.8290) by leveraging adaptive feature weighting, while sequence-based models such as H-LSTM, RNN, and GNN show relatively stable but limited generalization.

Notably, our proposed intention-driven BC--IRL framework achieves the highest F1-score (0.7733) and precision (0.7536) among all methods, while maintaining competitive accuracy. These results demonstrate the zero-shot generalization capability of our approach and confirm that the learned cooperation representation captures behaviorally meaningful patterns that transfer beyond the training dataset.

\subsection{Ablation Study}
Table~VI reports the ablation results to assess the contribution of each component in the proposed framework. Starting from the BC-only baseline, incorporating IRL leads to a consistent improvement across all metrics, indicating that reward regularization provides complementary supervision beyond pure imitation.
When enabling the cooperation module, introducing LCS or DCS individually does not yield immediate performance gains and may slightly reduce accuracy, suggesting that modeling only intrinsic or interaction-driven cooperation in isolation is insufficient. However, when LCS and DCS are jointly incorporated, the performance improves substantially (accuracy 0.9509), confirming that intrinsic and interactive cooperation modeling play complementary roles.

Finally, incorporating driving-style awareness maintains competitive performance and further stabilizes precision–recall balance. Although the version without explicit style modeling still performs strongly, the style-aware formulation achieves improved robustness and consistency across heterogeneous interaction patterns. Overall, these results demonstrate that structured cooperation modeling is critical to the effectiveness of the proposed BC–IRL framework.

\subsection{Cooperation-Aware Intention Analysis of Driving Styles}

Fig.~\ref{fig:coo_analysis} analyzes how cooperation-related components vary across driving styles.
As shown in Fig.~\ref{fig:coo_analysis}(a), the final cooperation score consistently increases from aggressive to conservative driving styles, with both the intrinsic cooperation score (LCS) and the interaction-based cooperation score (DCS) contributing more under conservative behavior. Notably, the relative contribution of DCS becomes more pronounced for conservative drivers, indicating a stronger reliance on surrounding vehicle interactions when making lane change decisions.
Fig.~\ref{fig:coo_analysis}(b) further reveals distinct cooperation profiles: aggressive drivers exhibit lower LCS and DCS values and assign a higher gating weight $\alpha$ to intrinsic preferences, reflecting a tendency to prioritize self-driven decisions. In contrast, conservative drivers show elevated LCS and DCS together with a reduced $\alpha$, implying that interaction-aware cooperation plays a more dominant role in shaping the final score.
These patterns suggest that the gating weight $\alpha$ adaptively mediates the balance between intrinsic intent and interaction awareness according to driving style, capturing the underlying behavioral differences in how drivers trade off self-preference and social cooperation during lane changes.

\subsection{Performance of Motion-planning Module}
Table~\ref{tab:lane_change_statistics} shows the quantitative results of lane change motion-planning experiments, including success rate (SR), LC duration, and jerk. By comparing with IDM method, our method get better performance in these metrics, demonstrating the significance of our method.
To qualitatively validate the motion-planning component, we analyze a representative lane change scenario and compare the proposed IRL–MPC approach with a traditional IDM baseline. Once a ``$LC$'' decision is made, the ego vehicle executes the maneuver while surrounding vehicles follow ground-truth trajectories from the NGSIM dataset. For comparison, the baseline ego vehicle behavior is generated by IDM, a widely used rule-based model in AV implementations.

\begin{figure*}[!t]
  \centering
  
  \subfloat[IRL-based trajectory prediction method.]{
    \includegraphics[width=0.48\textwidth]{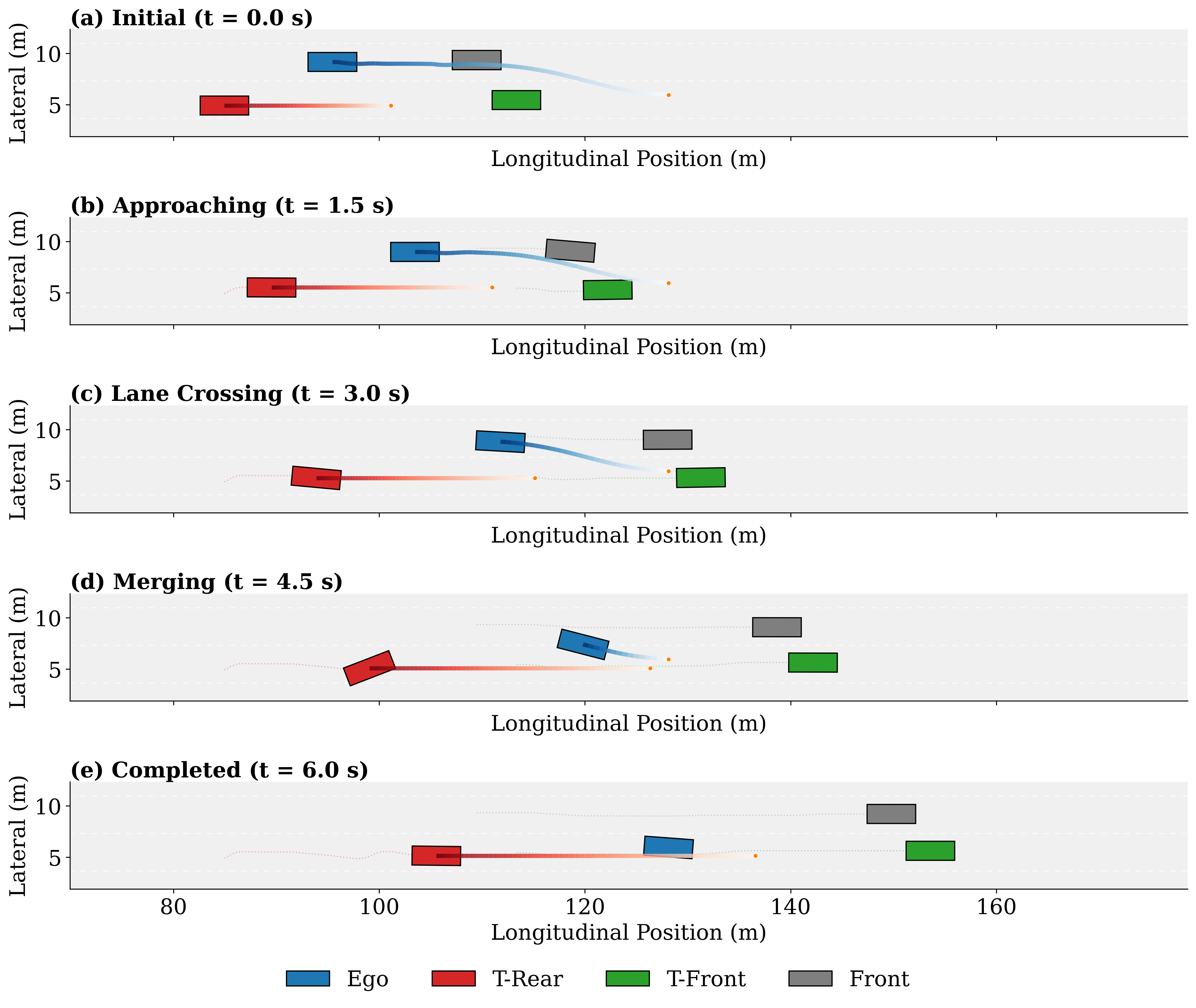}
    \label{fig:case_ours}
  }
  \hfill
  \subfloat[IDM-based trajectory prediction method.]{
    \includegraphics[width=0.48\textwidth]{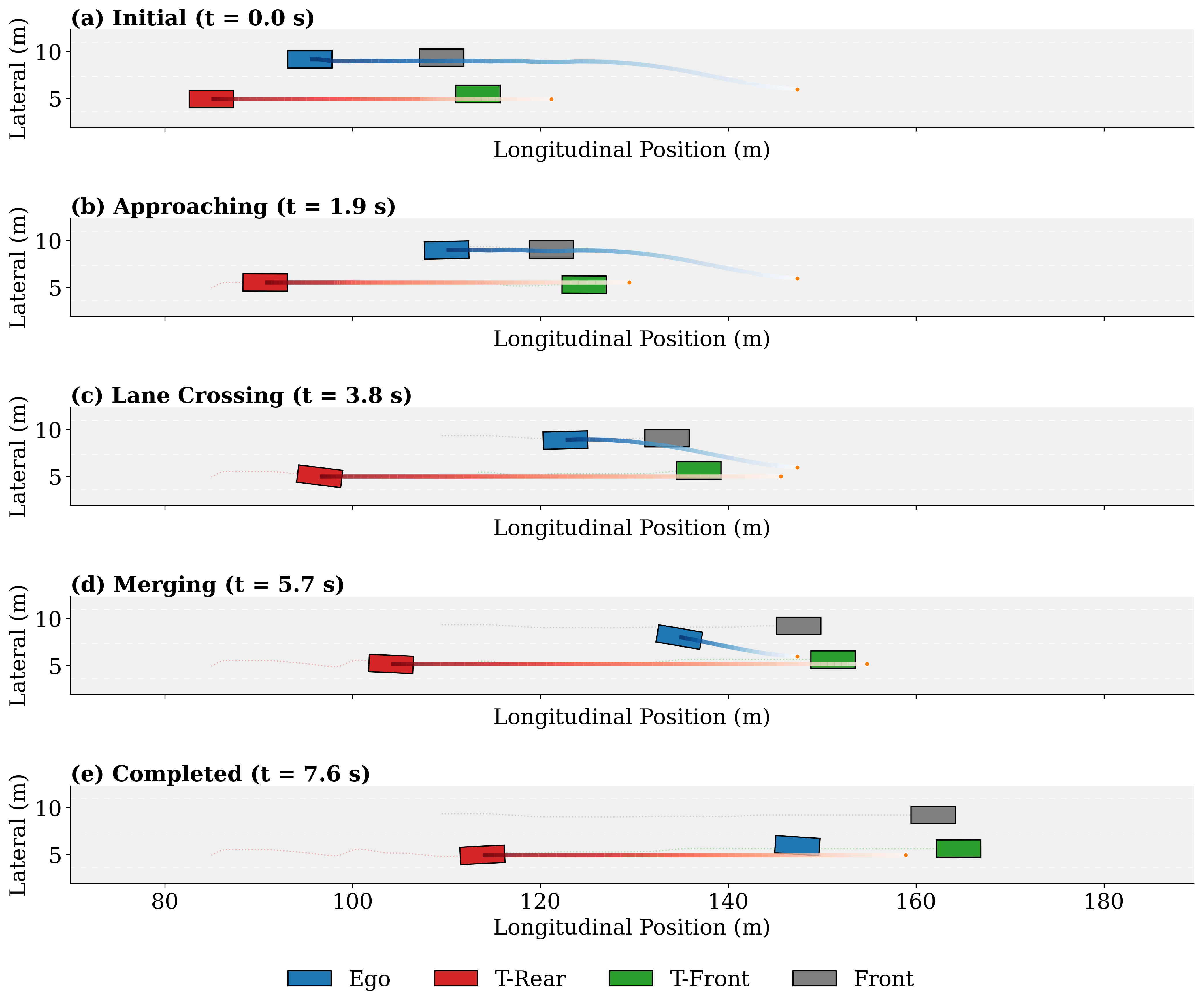}
    \label{fig:case_idm}
  }
  \caption{Qualitative results. Comparison of MPC-based lane change performance with predicted trajectories of the T-Rear vehicle. The blue vehicle represents the AV controlled by our driving framework. Other three vehicles symbolize the surrounding vehicles extracted from the NGSIM I-80 dataset. Blue line indicates the trajectory generated by the MPC, and the red line denotes the predicted trajectory of the T-Rear vehicle.}
  \label{fig:demo_mpc}
\end{figure*}

Fig.~\ref{fig:demo_mpc} shows prediction snapshots at one-second intervals (10 steps) for the T-Rear vehicle. The left column illustrates our IRL-based method, while the right column shows IDM predictions. The results demonstrate that our method enables the ego vehicle to complete the lane change substantially faster and more efficiently. Specifically, in Fig.~\ref{fig:demo_mpc}(a), the maneuver is completed within 6.0~s, whereas in Fig.~\ref{fig:demo_mpc}(b), IDM requires 7.6~s and only reaches mid-transition at 5.7~s.

By explicitly modeling driving style, our method achieves a higher success rate while reducing lane change duration, without sacrificing comfort or safety.
Beyond efficiency, the trajectories highlight differences in safety and adaptability. Under our approach, the ego vehicle initiates the lane change smoothly at a lower speed, maintaining a safe distance from the leading vehicle, and subsequently accelerates to finish the maneuver. In contrast, the IDM trajectory shows a delayed and slower transition, with higher potential safety risks due to less adaptive interaction with surrounding vehicles. These observations confirm that integrating reward-driven prediction and adaptive control yields lane changes that are not only more efficient but also safer and more human-like.

\begin{table}[t]
\centering
\small

\caption{Quantitative results of Lane Change Motion-Planning Experiments}
\label{tab:lane_change_statistics}
\begin{tabular}{lccccc}
\toprule
\multirow{2}{*}{Method} &
\multirow{2}{*}{SR$\uparrow$} &
\multicolumn{2}{c}{LC Duration$\downarrow$} &
\multicolumn{2}{c}{Jerk$\downarrow$} \\
& & Mean & Std & Mean & Std \\
& (\%) & (s) & (s) & (m/s$^3$) & (m/s$^3$) \\
\midrule
IDM           & 78.12 & 7.006 & 1.336 & 2.592 & 3.050  \\
Ours (w/o DS) & 83.75 & 6.674 & 1.453 & \textbf{2.409} & 2.726\\
\rowcolor[gray]{0.9}Ours (w/ DS)  & \textbf{86.83} & \textbf{6.548} & 1.462 & 2.459 & 2.644 \\
\bottomrule
\end{tabular}
\end{table}

\begin{figure}[!t]
  \centering
  \small
  \includegraphics[width=0.48\textwidth]{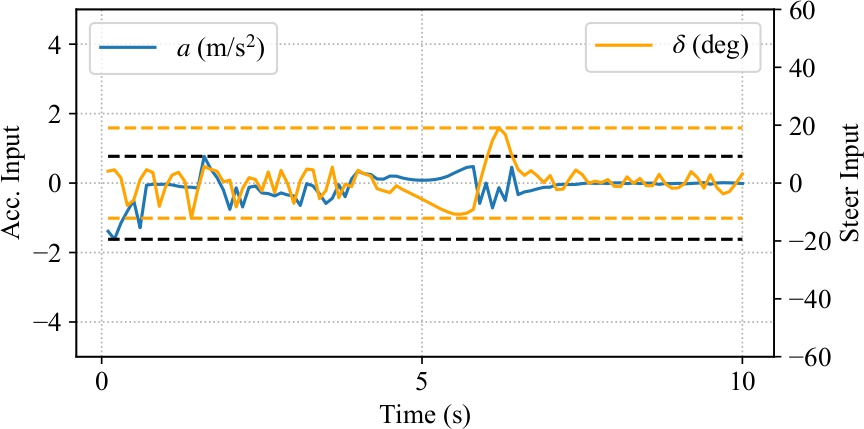}
  \caption{The control input for the ego vehicle when executing lane change maneuver.}\label{fig:mpc_control_input}
\end{figure}

\section{Discussions and Conclusion}
\subsection{Limitations and Future Work}
Despite the encouraging results, several limitations remain to be addressed in future work. First, while the current framework predicts the target vehicle’s trajectory and cooperation intention, it does not explicitly account for alternative maneuvers such as yielding, accelerating, or defensive lane changes, which may critically influence the ego vehicle’s planning. Second, our driving-style representation is restricted to three discrete categories, which simplifies heterogeneous human behaviors but overlooks finer-grained variations across drivers. Third, the current evaluation focuses on lane change scenarios, whereas real-world driving requires a broader spectrum of decisions, including car-following, merging, and gap acceptance in highly interactive contexts.  

Future research will therefore extend cooperation modeling by incorporating multi-intention inference for surrounding vehicles, allowing the ego vehicle to better anticipate reactive behaviors. We also plan to enrich the driving-style taxonomy by leveraging unsupervised or semi-supervised clustering to capture more nuanced behavioral patterns. Finally, we aim to develop a multi-agent cooperative decision-making framework that integrates both vehicle–vehicle and vehicle–infrastructure interactions, enabling AVs to operate safely and efficiently in diverse and congested traffic environments.

\subsection{Conclusion}
This paper presents an intention-driven lane change framework designed for mixed-traffic environments, where AVs must adapt to heterogeneous and unpredictable human driving behaviors. The framework synergistically integrates driving-style recognition, cooperation-aware intention modeling, decision-making, and motion-planning. Specifically, a driving-style classifier was built on the NGSIM dataset to enable real-time recognition of aggressive, normal, and conservative drivers. On this basis, we introduced intrinsic and interactive cooperation scores to quantify surrounding drivers’ cooperative intentions under different contexts. These scores were then incorporated into a BC–IRL decision-making model that balances supervised imitation with reward-driven generalization to determine when lane changes should be initiated. Finally, an IRL–MPC motion-planning module was employed to generate collision-free, socially compliant trajectories.  
Experimental evaluations demonstrate that the proposed framework achieves superior performance, with 96.98\% accuracy and 96.99\% F1-score, consistently outperforming other baselines. Cross-domain evaluation on the highD dataset further confirms its generalization capability. Notably, ablation studies verify the complementary roles of intrinsic and interaction-driven cooperation modeling, as well as the benefit of driving-style awareness. Beyond efficiency, the motion-planning results also highlight improvements in safety and comfort, indicating that cooperation-aware intention modeling facilitates more human-like decision-making and execution.

By explicitly modeling heterogeneous human intention and integrating it into both decision-making and motion-planning, this study provides a systematic framework for cooperation-aware lane change behavior in mixed-traffic environments. The incorporation of driving-style recognition, structured cooperation inference, and reward-guided planning enables more adaptive and interaction-consistent behavior of AVs. 
By bridging the gap between real-world human behaviors and automated lane change strategies, this work advances the development of context-aware and human-like AV systems for safe and efficient operations in complex traffic environments.

\section*{Acknowledgment}
This work was partially supported by NSFC Grant U24A20252 and 62373315, Guangdong provincial project 2023ZT10X009, Nansha Key Science and Technology Project under Grant 2023ZD006, and Guangzhou Bureau Of Education College Scientific Research Project 2024312169.

\bibliographystyle{IEEEtran} 
\bibliography{citations}

\section*{Appendix: Sensitivity Analysis on Reward Temperature}
\begin{table}[h]
\centering
\small

\caption{Sensitivity Analysis on Reward Temperature $\beta$.}
\label{tab:beta_sensitivity_appendix}
\begin{tabular}{lcccc}
\toprule
$\beta$ & Accuracy & Precision & Recall & F1 \\
\midrule
0 (no RW) & \textbf{0.9556} & \textbf{0.9561} & \textbf{0.9556} & \textbf{0.9558} \\
0.5       & 0.9547 & 0.9559 & 0.9547 & 0.9550 \\
1.0       & 0.9547 & 0.9552 & 0.9547 & 0.9549 \\
2.0       & 0.9528 & 0.9534 & 0.9528 & 0.9530 \\
4.0       & 0.9443 & 0.9460 & 0.9443 & 0.9448 \\
8.0       & 0.9330 & 0.9384 & 0.9330 & 0.9340 \\
16.0      & 0.9311 & 0.9328 & 0.9311 & 0.9316 \\
\bottomrule
\end{tabular}
\end{table}

\end{document}